\pdfoutput=1

\documentclass[11pt]{article}

\usepackage[preprint]{acl}

\usepackage{times}
\usepackage{latexsym}

\usepackage[T1]{fontenc}

\usepackage[utf8]{inputenc}

\usepackage{microtype}

\usepackage{inconsolata}

\usepackage{graphicx}

\usepackage{amsfonts}
\usepackage{amsmath}
\usepackage{multirow} 
\usepackage{array,multirow,diagbox}
\usepackage{colortbl}
\usepackage{bbding}
\usepackage{hwemoji}
\definecolor{myblue}{HTML}{2973B2}
\definecolor{mypurple}{HTML}{c0165f}

\usepackage{ntheorem}
\theoremseparator{:}

\title{\raisebox{-0.2cm}{\includegraphics[height=1.5em]{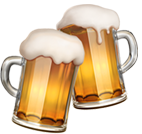}} Quaff: Quantized Parameter-Efficient Fine-Tuning under  Outlier Spatial Stability Hypothesis}

\author{Hong Huang \\
  Department of Computer Science \\
  City University of Hong Kong \\
  Hong Kong, China \\
  \texttt{honghuang2000@outlook.com} \\\And
  Dapeng Wu \\
  Department of Computer Science \\
  City University of Hong Kong \\
  Hong Kong, China \\
  \texttt{dpwu@ieee.org} \\}

\begin{document}

\maketitle

\begin{abstract}
Large language models (LLMs) have made exciting achievements across various domains, yet their deployment on resource-constrained personal devices remains hindered by the prohibitive computational and memory demands of task-specific fine-tuning.  While quantization offers a pathway to efficiency, existing methods struggle to balance performance and overhead, either incurring high computational/memory costs or failing to address activation outliers, a critical bottleneck in quantized fine-tuning.
To address these challenges, we propose the Outlier Spatial Stability Hypothesis (\textbf{OSSH}): \textit{During fine-tuning, certain activation outlier channels retain stable spatial positions across training iterations.} Building on OSSH, we propose \textbf{Quaff}, a Quantized parameter-efficient fine-tuning framework for LLMs, optimizing low-precision activation representations through targeted momentum scaling. Quaff dynamically suppresses outliers exclusively in invariant channels using lightweight operations, eliminating full-precision weight storage and global rescaling while reducing quantization errors. 
Extensive experiments across ten benchmarks validate OSSH and demonstrate Quaff's efficacy. Specifically, on the GPQA reasoning benchmark, Quaff achieves a $1.73\times$ latency reduction and $30\%$ memory savings over full-precision fine-tuning while improving accuracy by $0.6\%$ on the Phi-3 model, reconciling the triple trade-off between efficiency, performance, and deployability. By enabling consumer-grade GPU fine-tuning (\textit{e.g.,} RTX 2080 Super) without sacrificing model utility, Quaff democratizes personalized LLM deployment. The code is available at \url{https://github.com/Little0o0/Quaff.git}.
\end{abstract}

\section{Introduction}
Large language models (LLMs)~\citep{wu2023brief, floridi2020gpt, zhang2022opt} exhibit remarkable achievement in various domains, from creative writing to conversational chatbots. Growing demands for privacy-preserving, personalized LLMs (\textit{e.g.,} a virtual companion chatbot) deployed on local devices clash with the prohibitive computational and memory costs of fine-tuning, building a critical barrier for individuals and small enterprises.
While parameter-efficient fine-tuning (PEFT) reduces trainable parameters, it still imposes unsustainable overhead when scaling to billion-parameter models.

\begin{figure}[t]
\centering
\includegraphics[width=0.8\linewidth]{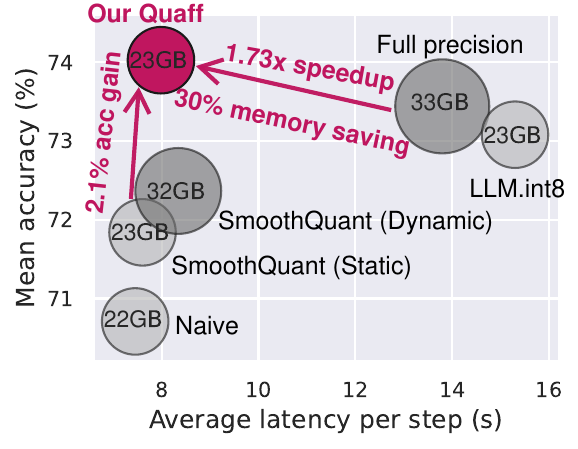}
\caption{Comparison of accuracy, average latency per step of WAQ baselines with Phi-3 on the GPQA benchmark using LoRA fine-tuning. The size of the marker represents the GPU memory footprints.}
\label{fig:firstimg}
\end{figure}
\begin{figure*}[t]
\centering
\includegraphics[width=\linewidth]{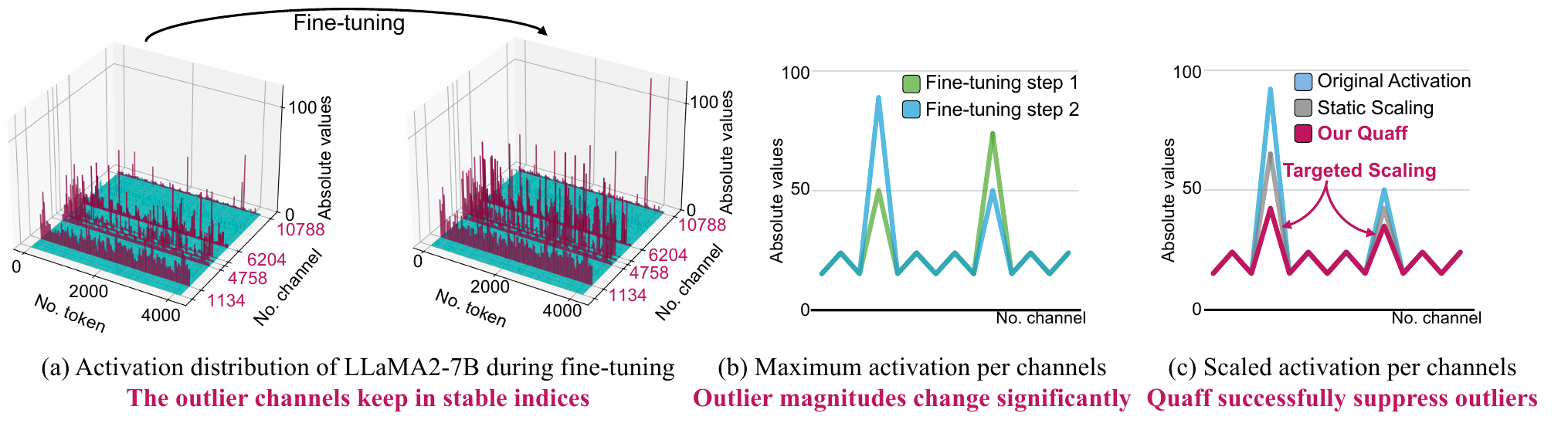}
\caption{ (a) Spatial stability of outlier channels: activation magnitude distribution during fine-tuning, demonstrating stable spatial indices for outlier channels across iterations. (b) Activation distribution shift: the magnitudes of outliers change significantly during fine-tuning. (c) Quaff Efficacy: Failure of static scaling due to outlier fluctuations, contrasted with Quaff's targeted momentum scaling on stable outlier channels, successfully suppressing outliers.}
\label{fig:activation}
\end{figure*}
Quantization~\citep{dettmers20218, dettmers2022llm, lin2023awq, frantar2022gptq} offers a pathway to efficiency, but most of existing work focuses narrowly on weight-only quantization (WOQ) in fine-tuning ~\cite{kwon2022alphatuning, dettmers2024qlora, xu2023qa, li2023loftq, liu2023qllm, guo2023lq, kim2024memory, he2023efficientdm, lee2024qeft, frantar2022gptq, lin2023awq}, which compresses frozen weights to low precision (\textit{e.g.,} 4-bit).  However, WOQ introduces computational bottlenecks via hardware-unfriendly mixed-precision operations between quantized weights and full-precision activations.
Weight-Activation Quantization (WAQ)~\cite{zhou2016dorefa, dettmers20218, hubara2018quantized} addresses this by quantizing both weights and activations into hardware-friendly formats (\textit{e.g.,} INT8), but faces a fundamental challenge in LLMs, \textbf{emergent channel-wise outliers}~\cite{wei2022outlier, wu2023understanding, yao2022zeroquant, dettmers2022llm, xiao2023smoothquant,huang2024rolora}, where large activations amplify quantization errors and degrade model performance.

Prior methods suppress outliers in \textit{inference} via channel-wise scaling between activations and weights~\cite{dettmers2022llm, wei2022outlier, xiao2023smoothquant, wei2023outlierplus} But these methods fail to adapt to \textit{fine-tuning} due to activation distribution shifts (Fig.~\ref{fig:activation} (b)): 1. Static scaling~\cite{xiao2023smoothquant} predefines factors on calibration data, but mismatched scaling amplifies quantization errors as distributions evolve. While some methods~\cite{lin2025duquant,huang2024rolora,ashkboos2024quarot} attempt to mitigate this by replacing scaling with rotation, they introduce architectural rigidity and computational Inefficiency. 2. Dynamic scaling~\cite{dettmers2022llm, xiao2023smoothquant} adapts factors in real time but requires storing and rescaling of full-precision weights, incurring prohibitive memory/compute costs.

We find the key problem in previous approaches is \textbf{coupling bottleneck}: scaled weight quantization depends on real-time activation statistics, preventing hardware-friendly deployment.
To address this, we first introduce the Outlier Spatial Stability Hypothesis (\textbf{OSSH}): \textit{During fine-tuning, certain activation outlier channels retain stable spatial position across training iterations.} Building on OSSH, we propose \textbf{Quaff}, a \textbf{Qua}ntized parameter-e\textbf{f}ficient \textbf{f}ine-tuning framework for LLMs that decouples weight and activation quantization via targeted momentum scaling. Quaff dynamically computing scaling factors exclusively for invariant outlier channels, eliminating full-precision weight storage and global rescaling, enabling low quantization error and high efficiency.

We evaluate Quaff across ten benchmarks, including reasoning (MMLU-Pro~\cite{wang2024mmlu}, GPQA~\cite{rein2023gpqa},  instruction-tuning(Alpaca-Finance~\cite{alpaca2023finance}, Self-instruct~\cite{wang2022self}) and long text task (LongForm~\cite{koksal2023longform}, LAMBADA~\cite{paperno2016lambada}), using LLaMA-2~\cite{touvron2023llama}, Phi-3~\cite{abdin2024phi}, and OPT~\cite{zhang2022opt} models with four general PEFT methods (LoRA\cite{hu2021lora}, Prompt tuning\cite{lester2021power}, P-tuning\cite{liu2021p} and IA3\cite{liu2022few}).
Our experiment also evaluates the deployability of Quaff on a consumer-grade laptop with RTX 2080 super GPU (8GB). 
Extensive experimental results suggest that Quaff achieves the best trade-offs: it achieves $1.73\times$ faster fine-tuning and $30\%$ memory reduction versus full-precision baselines while improving accuracy by $0.6\%$ on GPQA. Against state-of-the-art (SOTA) WAQ methods, Quaff delivers a $2.1\%$  accuracy gain under identical constraints (Fig.~\ref{fig:firstimg}), validating its ability to harmonize efficiency (latency/memory), accuracy, and deployability. \textit{To our knowledge, Quaff is the first work to systematically resolve the triple trade-off in WAQ LLM fine-tuning, enabling local-device fine-tuning as effortlessly as a toast.}

\section{Background and Challenge}
This section introduces foundational concepts of neural network quantization, analyzes the unique challenges of weight-activation quantization (WAQ) in LLMs, and identifies fundamental limitations in existing approaches for fine-tuning.
\subsection{Quantization Fundamentals}
Quantization reduces numerical precision in neural networks by converting weights and activations from high-bit formats (\textit{e.g.,} FP32) to low-bit representations (\textit{e.g.,} INT8), optimizing memory usage and computational efficiency. The standard symmetric round-to-nearest quantization~\cite{jacob2018quantization} maps a floating-point matrix $\mathbf{X}$ into an $N$-bit integer matrix $\mathbf{X}_{int}$: 
\begin{equation}
     \mathbf{X}_{int} = Q(\mathbf{X}) =\left[ \frac{\mathbf{X}}{\Delta_{\mathbf{X}}} \right], \quad \Delta_{\mathbf{X}} = \frac{\mathrm{max}(|\mathbf{X}|)}{2^{N-1} - 1},
    \label{eq:quant}
\end{equation}
where $[\cdot]$ denotes rounding function, and $\Delta_{\mathbf{X}}$ is quantization step size. The granularity of $\Delta_{\mathbf{X}}$ (scalar, vector, or matrix) determines how finely quantization is applied (see Appendix~\ref{sec:quant_gran}).

\subsection{Weight-Activation Quantization in LLMs}
Weight-activation quantization (WAQ) compresses both weights $\mathbf{W} \in \mathbb{R}^{c_{in}\times c_{out}}$ and activations $\mathbf{X} \in \mathbb{R}^{t\times c_{in}}$, where $t$ is the number of tokens, $c_{in}$ is the number of input channels, and $c_{out}$ is the number of output channels, to accelerate matrix multiplication $\mathbf{Y} = \mathbf{X}\mathbf{W}$:
\begin{equation}
    \mathbf{Y} \approx \Delta_{\mathbf{X}}\cdot(\mathbf{X}_{int}\mathbf{W}_{int})\cdot \Delta_{\mathbf{W}},
    \label{eq:quant_linear}
\end{equation}
where INT8 integer operations reduce compute costs by $4\times$ by the integer kernel. However, LLMs exhibit \textbf{emergent channel-wise outliers} (Fig.~\ref{fig:activation}) with magnitudes $100\times$ larger than typical activations~\cite{xiao2023smoothquant}, inflating $\Delta_{\mathbf{X}}$ and causing catastrophic quantization errors.

Prior work addresses outliers via channel-wise scaling~\cite{dettmers2022llm, wei2022outlier, xiao2023smoothquant, shao2023omniquant} by using:
\begin{equation}
\begin{split}
    \mathbf{Y} &= (\mathbf{X}\mathbf{s}^{-1})(\mathbf{s}\mathbf{W}) = \hat{\mathbf{X}}\hat{\mathbf{W}} \\
     &\approx \Delta_{\mathbf{\hat{X}}}\cdot(\mathbf{\hat{X}}_{int}\mathbf{\hat{W}}_{int})\cdot \Delta_{\mathbf{\hat{W}}},
    \label{eq:smooth}
\end{split}
\end{equation}
where $\hat{\mathbf{X}} =\mathbf{X}\mathbf{s}^{-1}$ denotes scaled activations, $\hat{\mathbf{W}}=\mathbf{s}\mathbf{W}$ denotes scaled weights\footnote{In this paper, we denote the multiplication between vector $\mathbf{s}$ and matrix $\mathbf{X}$ as element-wise, \textit{i.e.}, $[\mathbf{sW}]_{i,j} = \mathbf{s}_{i} \mathbf{W}_{i,j}$.}. The channel-wise factors $\mathbf{s} \in \mathbb{R}^{c_{in}}$ determined by both $\mathbf{W}$ and $\mathbf{X}$ suppresses outliers in activation $\mathbf{X}$ by redistributing them to $\mathbf{W}$. While effective for \textit{inference}, this approach falls short during \textit{fine-tuning}.

\subsection{Challenges}
Channel-wise scaling creates a  \textbf{coupling bottleneck} between weight and activation quantization: the scaled weights $\hat{\mathbf{W}} = \mathbf{sW}$ become dependent on real-time activations through scaling factors $\mathbf{s}$. 

This coupling manifests in two failure modes: 1.\textbf{ Static scaling} predefines $\mathbf{s}$ on calibration data and ignores activation distribution shifts during fine-tuning, leading to mismatched scaling factors (Fig.~\ref{fig:scaling}) that amplify quantization errors in $\mathbf{\hat{X}}$ and $\mathbf{\hat{W}}$. Some methods\cite{lin2025duquant,huang2024rolora,ashkboos2024quarot} attempt to address this by replacing scaling with rotation, setting $\mathbf{s}$ as a rotation matrix; however, they introduce computational inefficiency in computing $\mathbf{\hat{X}}$. 2. \textbf{Dynamic scaling} adapts $\mathbf{s}$  to activation distributions during fine-tuning, forcing repeated updating and requantization of $\hat{\mathbf{W}}$, requiring full-precision weight storage and dynamic recomputation, incurring unacceptable computational/memory costs. 

Existing methods~\cite{dettmers2022llm, wei2022outlier, xiao2023smoothquant, lin2025duquant} thus face a trilemma: preserving performance via dynamic scaling sacrifices efficiency; prioritizing efficiency and deployability via static scaling sacrifices adaptability. This fundamental limitation underscores the need for a decoupled quantization framework that preserves performance while enabling hardware-efficient fine-tuning.

\section{Methodology}
This section introduces the theoretical foundation of our approach, formalizes the Outlier Spatial Stability Hypothesis (OSSH), and presents the Quaff framework for efficient quantized fine-tuning.

\subsection{Motivation: Decoupling WAQ}
The core limitation of prior methods lies in the interdependence between scaled weights ($\mathbf{\hat{W}=\mathbf{sW}}$) and activation quantization.
To decouple this dependency, we reformulate channel-wise scaling in Eq.~\ref{eq:smooth} as:
\begin{equation}
\begin{split}
    \mathbf{Y} &= \mathbf{\hat{X}}(\mathbf{s}\mathbf{W}) = \mathbf{\hat{X}}\left(\mathbf{W} + (\mathbf{s}-\mathbf{1})\mathbf{W}\right) \\
&= \mathbf{\hat{X}}\underbrace{\mathbf{W}}_{\text{Static}} + \mathbf{\hat{X}}\underbrace{(\mathbf{s}-\mathbf{1})\mathbf{W}}_{\text{Dynamic}},
    \label{eq:qft1}
\end{split}
\end{equation}
isolating frozen, pre-quantizable weights $\mathbf{W}$ from the dynamic term $\mathbf{(s-1)W}$. Critically, for non-outlier channels $i$, activations do not need to be scaled, \textit{i.e.,} $\mathbf{s}_i = 1$,  rendering $(\mathbf{s - 1})$ highly sparse.
Letting $O$ denote the set of outlier channel indices, we simplify Eq.~\ref{eq:qft1} to:
\begin{equation}
\begin{split}
\mathbf{Y} &= \mathbf{\hat{X}}\mathbf{W} + \mathbf{\hat{X}}_{:,O}(\mathbf{s}_O-\mathbf{1})\mathbf{W}_O \\
 & = \mathbf{\hat{X}}\mathbf{W} + \mathbf{\hat{x}}\mathbf{\hat{w}},
\end{split}
\label{eq:qft2}
\end{equation}
where $\mathbf{\hat{x}} = \mathbf{\hat{X}}_{:,O}$ and $\mathbf{\hat{w}} = (\mathbf{s}_O-\mathbf{1})\mathbf{W}_O$ represent the submatrix of scaled activations and weights in outlier channels.\footnote{In this paper, we use the $\mathbf{X}_{:,O}$ to denote the submatrix containing $O$ columns of matrix $\mathbf{X}\in \mathbb{R}^{t\times c_{in}}$, \textit{i.e.,} $\mathbf{X}_{:,O} = (\mathbf{X}_{1,O},\dots,\mathbf{X}_{t,O})^T$, where $\mathbf{X}_{i,O} = (\mathbf{X}_{i,j}| j \in O)$.}
Notably, if outlier channels $O$ remain invariant during fine-tuning, storing only the small static submatrix $\mathbf{W}_O$ in full precision suffices to compute $\mathbf{\hat{w}}$ in real-time, avoiding costly dequantization of $Q(\mathbf{W})$, enabling hardware-efficient fine-tuning. Therefore, we propose OSSH to validate the invariant outlier channels.

\subsection{Outlier Spatial Stability Hypothesis}
Building on empirical observations of channel-wise outliers in LLM inference~\cite{dettmers2022llm, wei2022outlier, xiao2023smoothquant}, we enhance its spatial stability and extend it to the fine-tuning regime, formalizing this as the \textbf{Outlier Spatial Stability Hypothesis (OSSH)}: \\
\textit{During fine-tuning, certain activation outlier channels retain stable spatial positions across training iterations.}

Unlike prior methods~\cite{lee2024owq,xiao2023smoothquant} that assume activation stability based on empirical observations during inference (where the model is fixed), OSSH investigates outlier stability during fine-tuning, where activations undergo distributional shifts. This stability emerges from the interaction between (1) the preservation of foundational pre-trained features critical for cross-task generalization~\cite{mosbach2020stability} and (2) semantic consistency for salient tokens (e.g., [CLS])~\cite{fu2023stability,hewitt2019structural}. Outlier channels serve as anchors for high-level semantic primitives, bridging the model’s general knowledge and task-specific adaptations; their stability arises naturally because they consistently encode salient activations, ensuring robustness during fine-tuning.

\begin{figure}[t]
\centering
\includegraphics[width=0.8\linewidth]{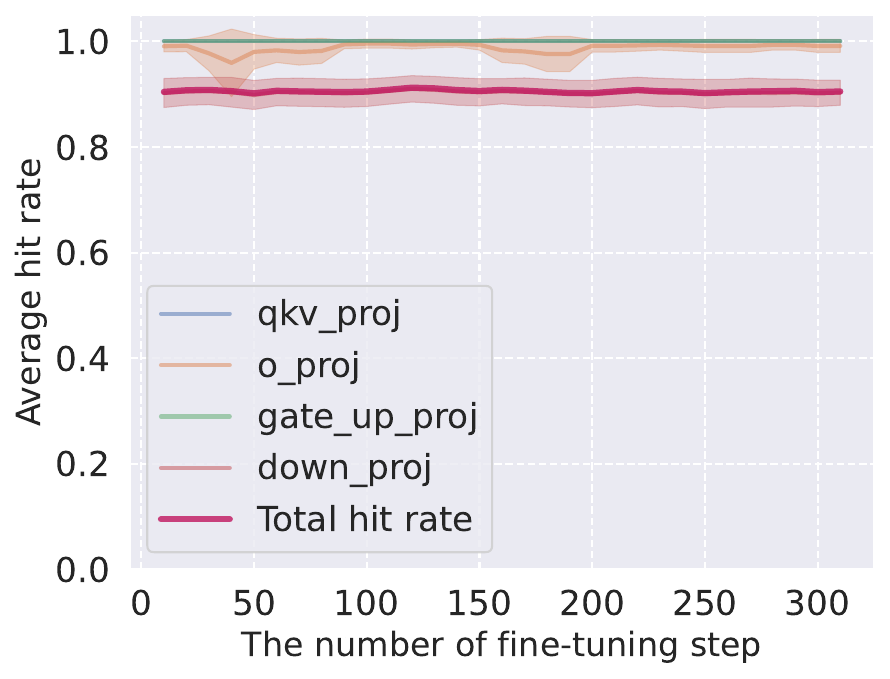}
\caption{The average hit rate of real-time versus predefined outlier channel indices across each layer of Phi3-3.8B during fine-tuning on OIG/Chip2. The shaded area represents the standard deviation.}
\label{fig:hitratephi}
\end{figure}

OSSH enables the pre-identification of static outlier channels $O$ prior to fine-tuning, eliminating resource-intensive runtime detection overhead. The validation in Fig.~\ref{fig:hitratephi} further supports the OSSH, where with $5\%$ predefined outlier channels, it can reach $>90\%$ overall hit rate across fine-tuning iterations. More analysis of OSSH is in Sec.~\ref{sec:OSSH}.

\subsection{Proposed Quaff}
Building on OSSH, we introduce a brand new quantized parameter-efficient fine-tuning algorithm, Quaff, decoupling the dependency between weight and activation quantization by targeted momentum scaling, where scaling factors are targeted computed in spatially invariant outlier channels with a momentum mechanism, optimizing efficiency without sacrificing performance.

Quaff begins with weights preprocessing before task-specific fine-tuning. First, Quaff uses a calibration dataset to identify outlier channels $O$. The outlier detection strategy can vary. For example, the criterion $\xi_o$ used to determine if a channel $o$ is an outlier channel can be defined as 
\begin{equation}
    \xi_o = \sum_{i}\mathbf{1}_{\max(|\mathbf{X}^{i}_{:,o}|) > 100\cdot\mathrm{mean(|\mathbf{X}^{i}|)}},
\end{equation}
where $\mathbf{X}^{i}$ denotes the activations from the $i$-th sample in the calibration datasets. Therefore, the outlier channel $O$ can be obtained by selecting outlier channels $o$ based on $\xi_{o}$. After that, Frozen weights $\mathbf{W}$ are quantized to $\mathbf{W}_{int}$ and $\Delta_{\mathbf{W}}$ as in Eq.~\ref{eq:quant},  while retaining full-precision of outlier channel weights $\mathbf{W}_{O}$. Based on empirical observations, we limit the overall overhead for $\mathbf{W}_{O}$ to less than $5\%$ by controlling the size of $O$. It should be noted that this 5\% budget is not uniformly distributed across all layers. As shown in previous work~\cite{lin2025duquant}, certain layers, such as $q\_proj$, contain few outlier channels, whereas others, like $down\_proj$, have a higher proportion of outlier channels. To accommodate this variance, we reallocate the budget from layers with fewer outliers, like $q\_proj$, to those with more, such as $down\_proj$, ensuring the total overhead remains below 5\%.

After weights preprocessing, Quaff injects learnable task-specific parameters $\theta$ (\textit{e.g.,} LoRA adapters) for fine-tuning. During the fine-tuning, a targeted momentum scaling mechanism is employed to stabilize outlier suppression. At the $t$-th step, scaling factors $\mathbf{s}_t$ blend historical values with current observations:
\begin{equation}
    \mathbf{s}_t = \gamma \mathbf{s}_{t-1} + (1 - \gamma)\beta ,
    \label{eq: momentum}
\end{equation}
where $\gamma \in [0, 1]$ is the hyperparameters to control update inertia, and the $\beta \in \mathbb{R}^{c_{in}}$ is defined as:
\begin{equation}
\beta_i = \begin{cases}
1, &\text{$i \notin O$}\\ 
\max\left(1, \sqrt{\frac{\max(|\mathbf{X}_{:,i}|)}{\max(|\mathbf{W}_i|)}} \right), &\text{$i \in O$}
\end{cases}.
\label{eq:def_s}
\end{equation}
This formulation prevents overreaction to transient activation shifts while maintaining compatibility with quantized weights. Then, the Quaff obtains the scaled activation
$\mathbf{\hat{X}}$ and scaled outlier weights $\mathbf{\hat{w}}$ by Eq.~\ref{eq:qft2}.
After that, using the uniform quantization as in Eq.~\ref{eq:quant},The forward pass in Eq.~\ref{eq:qft2} approximates:
\begin{equation}
\begin{split}
    \mathbf{Y} &\approx \Delta_{\mathbf{\hat{X}}}\mathbf{\hat{X}}_{int}\mathbf{W}_{int}\Delta_{\mathbf{W}} + \Delta_{\mathbf{\hat{x}}}\mathbf{\hat{x}}_{int}\mathbf{\hat{w}}_{int}\Delta_{\mathbf{\hat{w}}} \\
    & = \Delta_{\mathbf{\hat{X}}}(\mathbf{\hat{X}}_{int}\mathbf{W}_{int}\Delta_{\mathbf{W}} + \mathbf{\hat{x}}_{int}\mathbf{\hat{w}}_{int}\Delta_{\mathbf{\hat{w}}}),
\end{split}
\end{equation}
where $\Delta_{\mathbf{\hat{X}}} = \Delta_{\mathbf{\hat{x}}}$ and $\mathbf{\hat{x}}_{int}= [\mathbf{\hat{X}}_{int}]_{:,O}$ inherit activation quantization without overhead.

Compared to prior work, Quaff successfully addresses the trilemma of efficiency, performance, and deployability. 
\textbf{First}, Quaff reduces recomputation and memory overheads in scaling by $99\%$ compared to dynamic scaling methods~\cite{xiao2023smoothquant,dettmers2022llm} by targeted operations on the outlier channels $O$. Compared to naive WAQ in Eq.~\ref{eq:quant_linear}, Quaff incurs less than $5\%$ overhead (storing $\mathbf{W}_O$ and computing $\mathbf{\hat{x}}_{int}\mathbf{\hat{w}}_{int}\mathbf{\Delta}_{\mathbf{\hat{w}}}$), while significantly reducing quantization error through outlier suppression.
\textbf{Second}, the momentum-based scaling mechanism further stabilizes fine-tuning by smoothing out transient fluctuations in activation, prioritizing persistent distributional patterns for robust scaling. What's more, the $(\mathbf{s}-1)$ scaling terms reduce weight sensitivity relative to direct $\mathbf{s}$ scaling, enhancing quantization stability. 
\textbf{Lastly}, and most importantly, the efficiency of Quaff enables deployment of fine-tuning on edge devices (\textit{e.g.,} RTX 2080 Super) via a server-client paradigm: public servers preprocess and distribute quantized model weights $\mathbf{W}_{int}$ and outlier weights $\mathbf{W}_O$, while clients perform personalized quantized fine-tuning without storing full-precision weights.

\begin{figure*}[t]
\centering
\includegraphics[width=\textwidth]{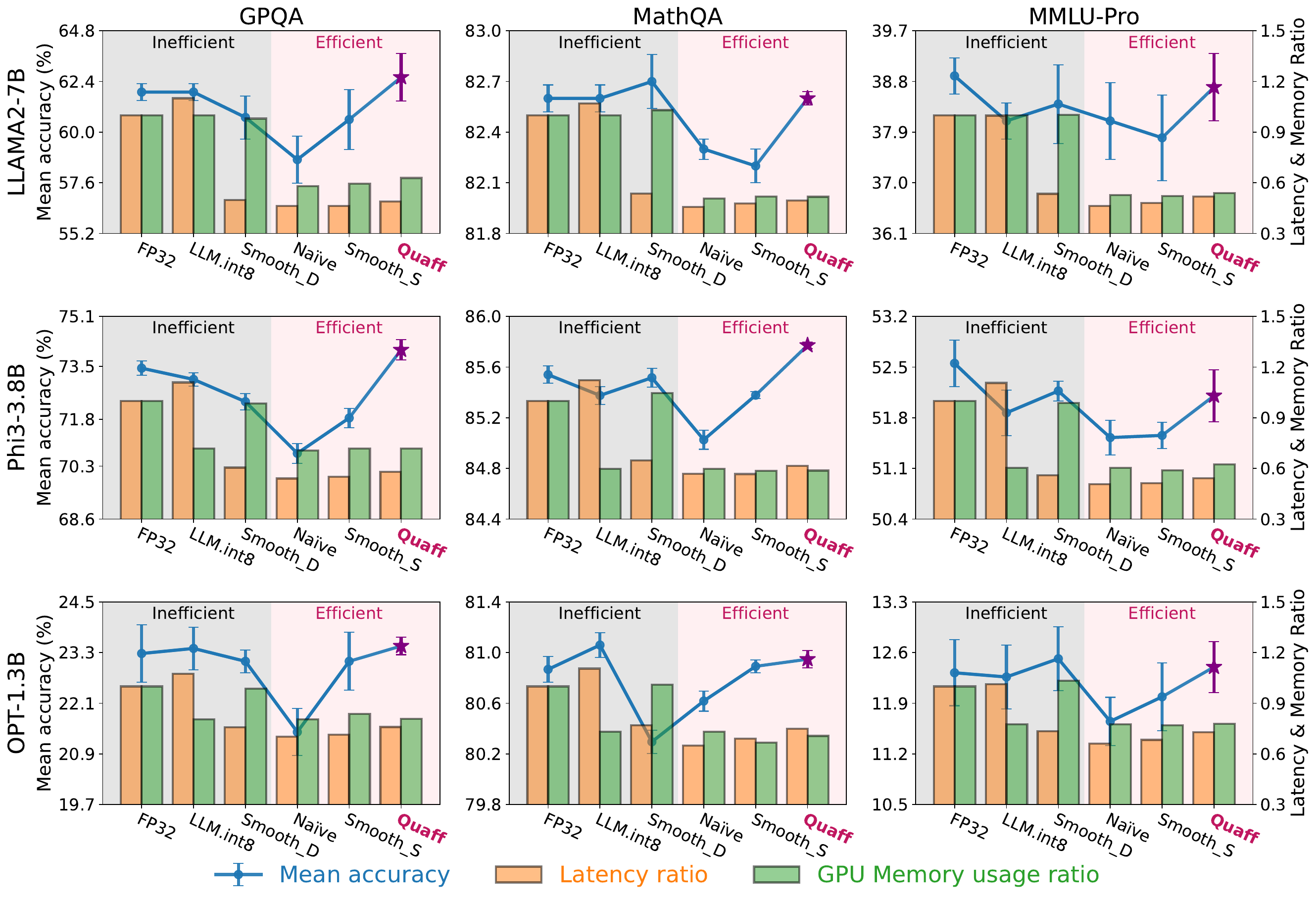}
\caption{Comparison of accuracy, latency, and memory footprints between our proposed Quaff and various WAQ baselines on three reasoning datasets using LoRA fine-tuning. Latency and memory footprint values are reported as ratios relative to FP32 models.}
\label{fig:main}
\end{figure*}
\begin{table*}[]
\resizebox{\textwidth}{!}{
\begin{tabular}{l|cc|cccccccccccc}
\hline
 & \multirow{2}{*}{\begin{tabular}[c]{@{}c@{}} \\ Latency \end{tabular}} & \multirow{2}{*}{\begin{tabular}[c]{@{}c@{}} \\ Memory\end{tabular}} & \multicolumn{3}{c|}{Oasst1} & \multicolumn{3}{c|}{Self-instruct} & \multicolumn{3}{c|}{Finance-Alpaca} & \multicolumn{3}{c}{HH-RLHF} \\
 &  &  & ROUGE-L$\uparrow$ & PPL$\downarrow$ & \multicolumn{1}{c|}{Acc$\uparrow$} & ROUGE-L$\uparrow$ & PPL$\downarrow$ & \multicolumn{1}{c|}{Acc$\uparrow$} & ROUGE-L$\uparrow$ & PPL$\downarrow$ & \multicolumn{1}{c|}{Acc$\uparrow$} & ROUGE-L$\uparrow$ & PPL$\downarrow$ & Acc$\uparrow$ \\ \hline
FP32 & 7.86s &  24.1GB  & 0.582 & 5.295 & \multicolumn{1}{c|}{0.679} & 0.670 & 2.086 & \multicolumn{1}{c|}{0.817} & 0.618  & 4.509 & \multicolumn{1}{c|}{0.608} & 0.511 & 4.579 & 0.605 \\ \hline
\rowcolor{gray!20} LLM.int8 & 8.92s &  16.4GB  & 0.573 & 5.519 & \multicolumn{1}{c|}{0.669} & {\color{myblue}\textbf{0.670}} & 2.103 & \multicolumn{1}{c|}{0.811} & 0.615 & \color{myblue}\textbf{4.599} & \multicolumn{1}{c|}{\color{mypurple}\textbf{0.607}} & 0.510 & \color{mypurple}\textbf{4.565} & 0.608 \\
\rowcolor{gray!20} Smooth\_D & 4.48s  &  23.0GB & 0.578 & \color{myblue}\textbf{5.323} & \multicolumn{1}{c|}{0.675} & 0.619  & 2.186 & \multicolumn{1}{c|}{0.820} & 0.614 & 4.605 & \multicolumn{1}{c|}{0.606} & 0.510 & 4.616 & 0.607 \\
\rowcolor{pink!20} Naive & 4.06s & 14.6GB  & 0.578 & 5.382 & \multicolumn{1}{c|}{\color{myblue}\textbf{0.676}} & 0.650 & 2.123 & \multicolumn{1}{c|}{0.820} & 0.615 & 4.633 & \multicolumn{1}{c|}{0.605} & 0.510 & 4.614 & 0.608 \\
\rowcolor{pink!20} Smooth\_S & 4.09s & 14.7GB  & \color{myblue}\textbf{0.579} & 5.342 & \multicolumn{1}{c|}{0.676} & 0.636 & \color{myblue}\textbf{2.099} & \multicolumn{1}{c|}{\color{myblue}\textbf{0.822}} & {\color{myblue}\textbf{0.616}} & 4.645 & \multicolumn{1}{c|}{0.605} & {\color{myblue}\textbf{0.511}} & 4.595 & \color{myblue}\textbf{0.610} \\
\rowcolor{pink!20} \color{mypurple}\textbf{Quaff} & 4.35s & 14.9GB  & \color{mypurple}\textbf{0.581} & \color{mypurple}\textbf{5.295} & \multicolumn{1}{c|}{\color{mypurple}\textbf{0.678}} & \color{mypurple}\textbf{0.682} & \color{mypurple}\textbf{2.098}  & \multicolumn{1}{c|}{\color{mypurple}\textbf{0.823}} & \color{mypurple}\textbf{0.617} & \color{mypurple}\textbf{4.595} & \multicolumn{1}{c|}{\color{myblue}\textbf{0.606}} & \color{mypurple}\textbf{0.512} & \color{myblue}\textbf{4.576} & \color{mypurple}\textbf{0.611} \\ \hline
\end{tabular}
}
\caption{ROUGE-L, Perplexity (PPL) and accuracy (Acc) on four instruction-tuning datasets with  Phi3-3.8B using LoRA fine-tuning. We report the average latency per step and the maximum GPU memory usage during fine-tuning. } 
\label{tb:ppl}
\end{table*}

\begin{figure*}[t]
\centering
\includegraphics[width=\linewidth]{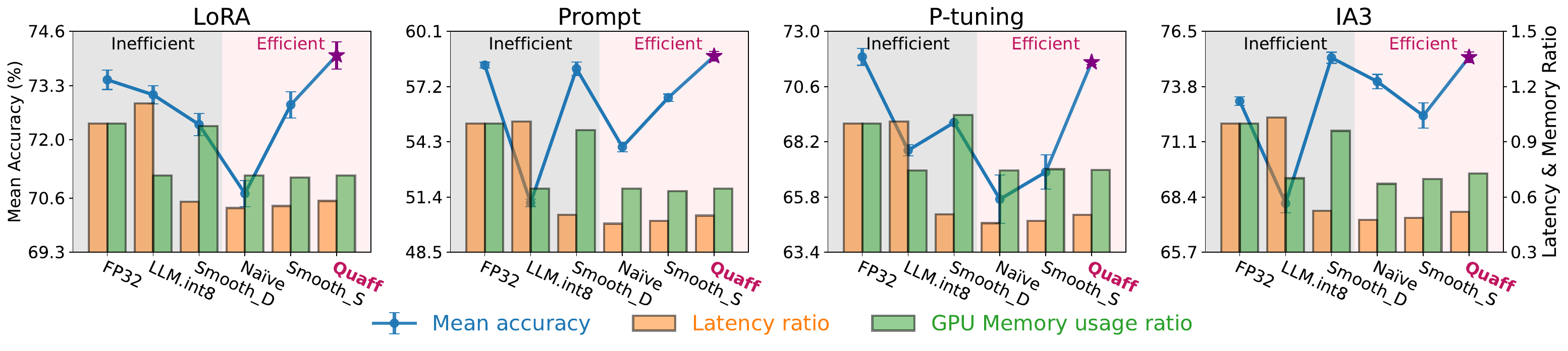}
\caption{Accuracy and fine-tuning costs on the GPQA dataset using Phi3-3.8B with different fine-tuning strategies.}
\label{fig:peft}
\end{figure*}

\section{Evaluation}
In this section, we comprehensively evaluate Quaff by comparing it against SOTA quantization approaches across ten benchmarks. All experiments are repeated five times, with means and standard deviations reported. In the table, the best and second-best results are highlighted in {\color{mypurple}purple} and {\color{myblue}blue}, respectively.
\subsection{Experimental Setup}
\paragraph{Datasets settings.} To ensure broad applicability, we evaluate the effectiveness of the proposed Quaff on diverse datasets spanning various downstream domains. Specifically, we use five instruction-tuning datasets (Alpaca-Finance~\cite{alpaca2023finance}, HH-RLHF~\cite{bai2022training}, Self-instruct~\cite{wang2022self}, OIG/Chip2~\cite{laion2023oig}, and Oasst1~\cite{kopf2024openassistant}), three reasoning datasets (GPQA~\cite{rein2023gpqa}, MathQA~\cite{amini2019mathqa}, and MMLU-Pro~\cite{wang2024mmlu}), and two long text datasets (Longform~\cite{koksal2023longform} and LAMBADA~\cite{paperno2016lambada}).

The selection of these datasets is motivated by the need for a comprehensive evaluation, particularly given that existing LLMs~\cite{touvron2023llama,abdin2024phi} and quantization methods~\cite{touvron2023llama, xiao2023smoothquant, dettmers2024qlora, dettmers2022llm} have primarily focused on these benchmarks. For datasets without predefined training and testing splits, we randomly sample 80\% of the data for training and allocate the remaining 20\% for testing. Details for benchmarks are in Sec.~\ref{sec:exp_setting}.

\paragraph{Model and Training settings.} We evaluate Quaff on LLaMA2~\cite{touvron2023llama}, Phi3~\cite{abdin2024phi}, and OPT~\cite{zhang2022opt} using four popular PEFT methods: LoRA~\cite{hu2021lora}, Prompt~\cite{lester2021power}, P-tuning~\cite{liu2021p} and IA3~\cite{liu2022few}. The default fine-tuning batch size is set to 16. We set a quantization precision of INT8 to ensure broad hardware compatibility. The models are maintained at 32-bit floating-point (FP32) precision by default. We opted for FP32 instead of FP16 because many LLMs (\textit{e.g.,} LLaMA\cite{paperno2016lambada}) only support Brain Floating Point 16-bit (BF16) training rather than FP16, and BF16 requires specialized, advanced GPUs (at least with the Ampere architecture), limiting accessibility on many devices.

For outlier channel identification, we use 512 data samples from OIG/Chip2~\cite{laion2023oig} as the calibration dataset. For different layers, we set the size of $O$ differently. Specifically, we allocate a maximum budget of $0.03\%c_{in}$ outlier channels for the $q\_proj$, $k\_proj$, $v\_proj$, $up\_proj$, $4\%c_{in}$ for $o\_proj$ , and $10\%c_{in}$ for $down\_proj$. The overall maximum overhead for outlier channels is maintained at less than $5\%$. Moreover, to simulate a typical user scenario, most experiments are conducted on a mid-range GPU, RTX 5880 Ada, which offers a similar computing speed to an RTX 4080 but with higher GPU memory. Additionally, some experiments are performed on a laptop with an RTX 2080 Super to demonstrate performance under custom-grade devices. Detailed hyperparameters for the experimental settings are provided in Sec.~\ref{sec:exp_setting}.

\paragraph{Baseline settings.} We compare our proposed Quaff framework with naive quantization (Naive), LLM.int8~\cite{dettmers2022llm}, and SmoothQuant~\cite{xiao2023smoothquant}, which has two versions: static (Smooth\_S) and dynamic (Smooth\_D).  We also include the (FP32) in our experiments. We also include the FP32 baseline in our experiments. We followed the settings in the paper of baselines. Certain rotation-based WAQ methods, such as DuQuant~\cite{lin2025duquant} and RoLoRA~\cite{huang2024rolora}, are excluded from our comparison due to their computational inefficiency and architectural rigidity in rotation during fine-tuning. Further discussion on the baseline methods is in Sec.~\ref{sec:baseline}.

\subsection{Performance Evaluation}
In the result, methods are categorized into efficient and inefficient approaches based on latency and memory footprint levels, with {\color{pink}pink} and {\color{gray}gray} backgrounds, respectively.
\paragraph{Fine-tuning on Reasoning Tasks.}
To demonstrate the efficiency and effectiveness of Quaff, we compare it with other WAQ baselines on three reasoning benchmarks (GPQA, MMLU-Pro, and MathQA), using OPT-1.3B, Phi3-3.8B, and LLaMa2-7B models with LoRA fine-tuning. 
A comprehensive comparison in terms of performance, end-to-end latency, and maximum GPU memory usage during fine-tuning is presented in Fig.~\ref{fig:main}. 
Remarkably, Quaff achieves the best trade-off among accuracy, computational cost, and memory footprint across all cases.
For instance, on the GPQA dataset using the LLaMA2-7B model, Quaff delivers a 2.0\% accuracy gain over comparable baselines with similar efficiency. Additionally, Quaff reduces latency by 51.1\% and GPU memory usage by 37.1\% versus FP32, even slightly outperforming full-precision training. These results validate Quaff’s ability to mitigate activation outlier impacts without sacrificing performance. Moreover, Quaff achieves better or at least similar performance compared to Smooth\_D, which dynamically scales all channels, demonstrating outlier channel invariance as assumed in the OSSH.

Notably, the Phi3-3.8B model surpasses the LLaMA2-7B model in accuracy across most datasets despite having only half the parameters.  Given this empirical trend, we select the Phi3-3.8B as our default model for subsequent experiments.

\begin{table}[]
\resizebox{\linewidth}{!}{
\begin{tabular}{l|cc|ccc}
\hline
 & Latency & Memory & ROUGE-L$\uparrow$  &  PPL$\downarrow$ & Acc$\uparrow$  \\ \hline
FP32 & 115.76s & 8+7.1G & 0.598 & 4.042
 & 0.665 \\ \hline
\rowcolor{gray!20} Smooth\_D & 131.67s & 8+7.1G & 0.589 & 4.116 & 0.663 \\
\rowcolor{pink!20} LLM.int8 & 20.43s  &  6.6G & 0.633  & 3.185 & 0.697 \\
\rowcolor{pink!20} Naive & 10.90s & 6.3G  & {\color{myblue}\textbf{0.639}} & 2.995 & 0.705 \\
\rowcolor{pink!20} Smooth\_S & 11.90s  &  6.3G & 0.638  & {\color{myblue}\textbf{2.970}} & {\color{myblue}\textbf{0.706}} \\
\rowcolor{pink!20} \color{mypurple}\textbf{Quaff} & 12.46s & 6.3G & {\color{mypurple}\textbf{0.643}} & {\color{mypurple}\textbf{2.962}} &  {\color{mypurple}\textbf{0.707}} \\ \hline
\end{tabular}
}
\caption{Results after 24 hours of LoRA fine-tuning on the OIG/CHIP2 dataset using Phi3-3.8B on a laptop (RTX 2080 Super 8GB) with 16GB shared memory.}
\label{tab:laptop}
\end{table}
\paragraph{Fine-Tuning on Instruction-tuning Tasks.} To assess personalized chatbot adaptation, we evaluate Quaff on four instruction-tuning datasets (Oasst1, Self-Instruct, Finance-Alpaca, and HH-RLHF) using the Phi3-3.8B model. We report metrics including ROUGE-L, perplexity, average accuracy, average latency per fine-tuning step, and maximum GPU memory usage. As shown in Table~\ref{tb:ppl}, Quaff achieves the best/second-best perplexity, accuracy, and ROUGE-L scores across all tasks while maintaining low latency and memory usage. This demonstrates its viability for real-world conversational LLM deployment on local devices.
\begin{figure*}[t]
\centering
\includegraphics[width=\linewidth]{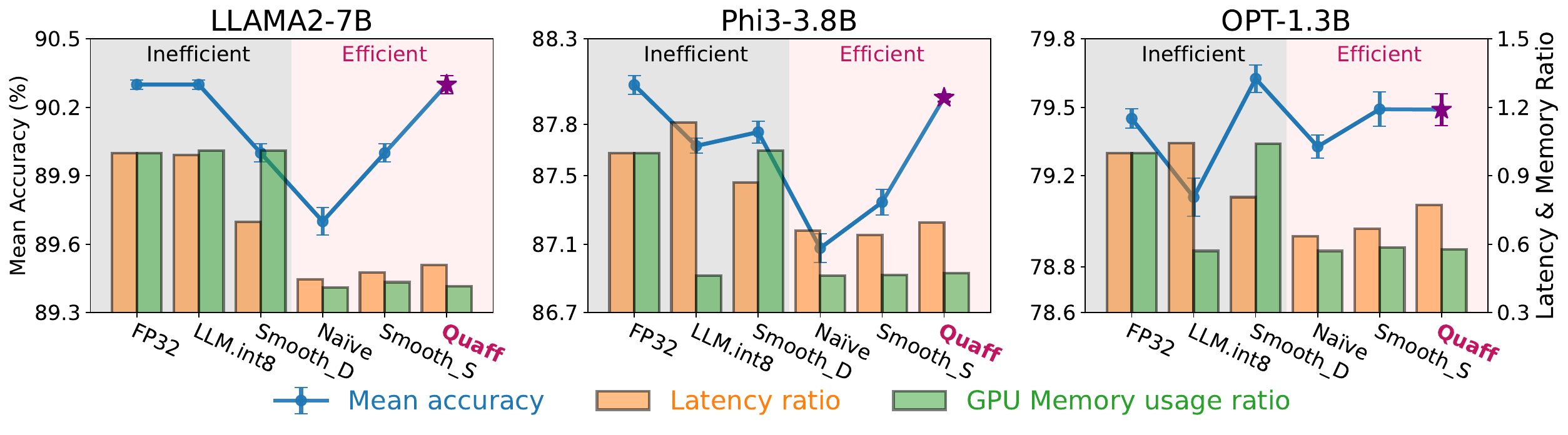}
\caption{Results of LoRA fine-tuning on LAMBADA dataset with input/output size of 4K on different models.}
\label{fig:lambada}
\end{figure*}

\paragraph{Consumer Hardware Compatibility.}
To emphasize accessibility for ordinary users, we evaluated the efficiency of Quaff and other baselines on an MSI laptop equipped with an NVIDIA RTX 2080 Super (8 GB) and an Intel Core i7 processor. We conducted experiments on the OIG/CHIP2 dataset using the Phi3-3.8B model for 24 hours of training, with a batch size of 1 and a gradient accumulation of 16. The results, illustrated in Tab.~\ref{tab:laptop}, show that Quaff achieves the best performance among all metrics. Due to out of CUDA memory, the FP32 and Smooth\_D suffer from high latency, so Quaff achieves $8.29 \times$ speedup versus FP32.

\paragraph{Fine-Tuning with Different Strategies.} 
Fine-tuning strategies may vary depending on the downstream task. To demonstrate the versatility of our proposed Quaff, we compare it with other WQA baselines on the GPQA dataset using four popular PEFT strategies (LoRA, Prompt, P-tuning, and IA3) with Phi3-3.8B model. Figure~\ref{fig:peft} shows Quaff outperforms all baselines, including FP32, across all strategies. This highlights the robustness of Quaff in diverse fine-tuning paradigms.

\paragraph{The Impact of Cross-Datasets Calibration.}
Different calibration datasets may produce distinct outlier channels, which may not be suitable for downstream tasks if the calibration dataset is mismatched. To investigate the selection of the calibration dataset, we evaluate Quaff on three datasets: OIG/CHIP2 (instruction-tuning), LAMBADA (long-context task), and GPQA (reasoning), using each as the calibration dataset for the others. We track ROUGE-L for OIG/CHIP2, accuracy for LAMBADA and GPQA. The results shown in Tab.~\ref{tb:cross} indicate that OIG/CHIP2 performs best as the calibration dataset, highlighting the advantage of instruction-tuning datasets for calibration.

\begin{table}[]
\resizebox{\linewidth}{!}{
\begin{tabular}{l|cccc} \hline 
&  OIG/CHIP2 & LAMBADA & GPQA \\  \hline 
 OIG/CHIP2 & 0.665 & 0.880 & 0.740  \\
LAMBADA & 0.659 & 0.880 &  0.733  \\
GPQA & 0.660 & 0.873 & 0.742 \\
\hline 
\end{tabular}
}
\caption{The Impact of cross-dataset calibration on Rouge-L (OIG/CHIP2) and accuracy (others). Columns: fine-tuning datasets; rows: calibration datasets.}
\label{tb:cross}
\end{table}

\paragraph{Fine-Tuning on Long Text Tasks.} 
In long-text tasks, activations in large language models (LLMs) become increasingly complex and unpredictable. To demonstrate that our Quaff approach maintains both effectiveness and efficiency in such scenarios, we conducted experiments on the LAMBADA~\cite{paperno2016lambada} (last word prediction) and LongForm~\cite{koksal2023longform} datasets (instruction following), designed for long-context understanding and long-text generation, respectively.  Both input and output sequences were set to a maximum length of 4K tokens, with a batch size of 1 and gradient accumulation of 16. Quaff achieved performance comparable to FP32 models in long-context language understanding (Fig.~\ref{fig:lambada}) and the best/second-best performance among WAQ methods in the long-text generation task(in Tab.~\ref{tb:longform}).

\begin{table}[]
\resizebox{\linewidth}{!}{
\begin{tabular}{l|cccc} \hline 
 & LoRA & Prompt & P-Tuning & IA3 \\  \hline 
Best baseline & 0.731 & 0.581 & 0.690 & {\color{myblue}\textbf{0.751}} \\
Quaff w/o Mo & {\color{myblue}\textbf{0.732}} & {\color{myblue}\textbf{0.583}} & {\color{myblue}\textbf{0.710}} & 0.745 \\
Quaff & {\color{mypurple}\textbf{0.740}} & {\color{mypurple}\textbf{0.588}} & {\color{mypurple}\textbf{0.717}} & {\color{mypurple}\textbf{0.752}} \\
\hline 
\end{tabular}
}
\caption{Mean accuracy on the GPQA dataset using the Phi3-3.8B model for Quaff, Quaff without Momentum, and the best baseline across different fine-tuning strategies, where the best baseline refers to the highest results achieved among prior WAQ methods.}
\label{tb:momentum}
\end{table}

\paragraph{The Impact of Momentum Mechanism.} 
To evaluate the effectiveness of the momentum mechanism in Quaff, we conducted an ablation experiment isolating the impact of Quaff’s momentum mechanism on the GPQA dataset using the Phi3-3.8B model. The results, presented in Table~\ref{tb:momentum}, demonstrate that momentum-based scaling factors enhance Quaff’s performance, and improve accuracy by $0.5\%-0.8\%$ by prioritizing persistent Notably, even without the momentum mechanism, Quaff still outperforms the best baseline, highlighting the effectiveness of reduced weight sensitivity by $(\mathrm{s} - 1)$ scaling.

\begin{table}[t]
\resizebox{\linewidth}{!}{
\begin{tabular}{l|cc|ccc}
\hline
 & Latency & Memory & ROUGE-L$\uparrow$  & PPL$\downarrow$ & Acc$\uparrow$  \\ \hline
FP32 & 14.14s & 27.0GB & 0.516 & 7.865 & 0.594 \\ \hline
\rowcolor{gray!20} LLM.int8 & 16.36s  & 18.0GB & 0.511 & {\color{mypurple}\textbf{8.068}} & 0.589 \\
\rowcolor{gray!20} Smooth\_D & 15.79s & 27.6GB & {\color{myblue}\textbf{0.512}} & 8.199 & {\color{myblue}\textbf{0.590}} \\
\rowcolor{pink!20} Naive & 10.28s & 17.6GB & 0.510 & 8.262 & 0.588 \\
\rowcolor{pink!20} Smooth\_S & 10.53s  &  17.7GB &  0.510 & 8.262 & 0.589 \\
\rowcolor{pink!20} \color{mypurple}\textbf{Quaff} & 11.60s & 17.7GB & {\color{mypurple}\textbf{0.513}} & {\color{myblue}\textbf{8.189}} &  {\color{mypurple}\textbf{0.590}} \\ \hline
\end{tabular}
}
\caption{Results of LoRA fine-tuning on the LongForm dataset with output size of 4K on Phi3-3.8B model.}
\label{tb:longform}
\end{table}
\section{Related Work}
\paragraph{Parameter-Efficient Fine-Tuning.}
Parameter-Efficient Fine-Tuning (PEFT) adapts tasks by training only a small subset of a pretrained model’s parameters. Techniques such as adapter tuning~\cite{houlsby2019parameter} (inserting lightweight modules), prefix/prompt tuning~\cite{li2021prefix,lester2021power} (prepending learnable tokens), LoRA~\cite{hu2021lora} (low-rank weight updates), and IA3~\cite{liu2022few} (scaling activations) reduce computational and memory costs versus full fine-tuning. Yet, for billion-parameter LLMs, these methods can still be too resource-intensive for edge-device deployment.

\paragraph{Quantization Fine-Tuning for LLMs.}
Quantization~\cite{jacob2018quantization} addresses the limitations of PEFT by compressing model weights and activations. Weight-Only Quantization (WOQ)~\cite{kwon2022alphatuning, dettmers2024qlora, xu2023qa, li2023loftq, liu2023qllm, guo2023lq, kim2024memory, he2023efficientdm, lee2024qeft} compresses pretrained weights to low precision~\cite{frantar2022gptq, lin2023awq} while retaining a small set of trainable parameters in full precision for updating. Although WOQ reduces memory usage, it introduces mixed-precision computational overhead, often resulting in slower training compared to full-precision baselines.

Weight-Activation Quantization (WAQ) quantizes both weights and activations for hardware-friendly computation~\cite{zhou2016dorefa}. However, LLMs exhibit emergent channel-wise outliers inflating quantization errors~\cite{wu2023understanding}. To mitigate this, previous works~\cite{dettmers2022llm, wei2022outlier, xiao2023smoothquant, wei2023outlierplus, wang2024outliertune} redistribute outliers to weights via channel-wise scaling. Some variants~\cite{ashkboos2024quarot, lin2025duquant, huang2024rolora, liu2024spinquant, kampeas2023rotation} refine this by replacing scaling with rotation, but they suffer from computational inefficiency and architectural rigidity for activation rotation during fine-tuning, as detailed in Sec.~\ref{sec:baseline}. Therefore, existing WAQ methods suffer from \textbf{coupling} between weight and activation quantization, and suffer from either high memory/compute overhead for handling full-precision weights (dynamic scaling) or accuracy loss from distribution shifts (static scaling). Quaff solves this by targeted scaling based on OSSH to decouple quantization, enabling efficient fine-tuning with near-FP32 accuracy and minimal overhead.

\section{Conclusion}
This paper proposes an Outlier Spatial Stability Hypothesis (OSSH): During fine-tuning, certain activation outlier channels
retain stable spatial positions across training iterations. Based on OSSH, we propose Quaff, a quantized parameter-efficient fine-tuning framework for LLMs. Quaff decouples the quantization between weights and activations by targeted momentum scaling on stable outlier channels, achieving lower quantization error with little overhead. We conduct extensive experiments on ten benchmarks and show that Quaff outperforms existing approaches in terms of performance, computational cost, and memory footprints. 

\section*{Acknowledgment}
This paper is partially supported by Hong Kong Research Grants Council (RGC) grant \#11203523.

\section*{Limitations}
Our work prioritizes democratizing quantized fine-tuning for non-expert users. Therefore, our work lacks in-depth exploration and design in the following areas: 
1. Larger Model. We focus exclusively on models up to 7B parameters (e.g., LLaMA-2-7B, Phi-3-3.8B), and do not consider larger, state-of-the-art models.
2. Hardware-Specific Optimizations. We did not use hardware-specific optimizations (such as Ampere Tensor Cores or H100 FP8 acceleration~\cite{kim2025investigation}) and top-tier GPUs (e.g., A100) in our experiments, limiting exploration of high efficiency on advanced hardware for enterprise users.
3. Precision Constraints. We adopt only INT8 quantization and do not explore INT4/INT2 precision, resulting in a lower compression rate.
4. Layer-Agnostic Implementation. Quaff applies uniform quantization to linear layers without custom fusion or exploiting sparsity, sacrificing potential latency gains from architecture-specific tuning. 
5. Single-GPU Focus. We only consider single-GPU scenarios in our experiments and do not explore multi-GPU configurations. 

These limitations reflect our commitment to accessibility and compatibility for non-expert users while acknowledging areas for future enhancement.

\newpage
\bibliography{acl_latex}
\newpage
\appendix
\section{More WAQ Method Analysis}
\label{sec:baseline}
We analyze key weight-activation quantization (WAQ) baselines beyond the classical scaling method~\cite{xiao2023smoothquant}, highlighting their computational and practical limitations compared to Quaff.

\paragraph{LLM.int8.}
LLM.int8~\cite{dettmers2022llm} employs mixed-precision outlier handling by dynamically detecting high-magnitude activation channels via a fixed threshold $\sigma$. During fine-tuning, it splits computations into:
\begin{equation}
\begin{split}
\mathbf{Y} &= \mathbf{XW} = \underbrace{\mathbf{X}_{:,\bar{O}}\mathbf{W}_{\bar{O}}}_{\text{Quantized}} + \underbrace{\mathbf{X}_{:,O}\mathbf{W}_{O}}_{\text{Full-Precision}},
\end{split}
\label{eq:llmint8 }
\end{equation}
where $O$ denotes outlier channels and $\bar{O}$ denotes normal channels. Though structurally similar to Quaff, LLM.int8’s reliance on dynamic detection forces full-weight dequantization to retrieve $W_{O}$, incurring prohibitive latency. As activation distributions shift during fine-tuning, $\mathrm{card}(O)$ often grows to match $c_{in}$ (Fig.~\ref{fig:main}), rendering memory savings negligible versus FP32.

Conceptually, LLM.int8 can be reframed as a dynamic scaling variant:
\begin{equation}
\begin{split}
    \mathbf{Y} &= \mathbf{XW} = \mathbf{Xs}^{-1}\mathbf{sW}+  \mathbf{X\bar{s}}^{-1}\mathbf{\bar{s}W}\\
    & = \mathbf{\hat{X}}\mathbf{\hat{W}}+  \mathbf{X\bar{s}}^{-1}\mathbf{\bar{s}W} \\
    & \approx \Delta_{\mathbf{\hat{X}}}\cdot(\mathbf{\hat{X}}_{int}\mathbf{\hat{W}}_{int})\cdot \Delta_{\mathbf{\hat{W}}} +  \mathbf{X\bar{s}}^{-1}\mathbf{\bar{s}W},
\end{split}
    \label{eq:llmint8}
\end{equation}
where $\mathbf{s} + \bar{\mathbf{s}} = \mathbf{1}$, and $\mathbf{s}_i = \mathbf{1}_{\max(|\mathbf{X_{:,i}}|>\sigma)}$, where $\sigma$ is a predefined threshold. This exposes its core inefficiency: real-time scaling factor computation and global requantization.

\paragraph{Rotation-Based Methods.}
Several rotation-based approaches~\cite{lin2025duquant, huang2024rolora, ashkboos2024quarot} also recognize that static scaling cannot effectively suppress outliers due to fluctuation. To address this, these methods replace scaling with orthogonal transformations:
\begin{equation}
\begin{split}
    \mathbf{Y} & = \mathbf{(XR)(R^{T}W)} = \mathbf{\hat{X}\hat{W}} \\
    & \approx \Delta_{\mathbf{\hat{X}}}\cdot(\mathbf{\hat{X}}_{int}\mathbf{\hat{W}}_{int})\Delta_{\mathbf{\hat{W}}},
\end{split}
\end{equation}

where $\mathbf{R}$ is an orthogonal matrix satisfying $\mathbf{RR}^{T} = \mathbf{I}$ and $\mathbf{|R|} = 1 $. While effective for reducing post-training quantization error, rotation-based methods incur significant computational overhead. Although the Fast Walsh-Hadamard Transform (FWHT) reduces rotation complexity to $O(n log n)$ complexity, its recursive computation pattern limits vectorization and parallelism on general hardware and increases memory access fragmentation, making it much less efficient than hardware-friendly scaling methods. To mitigate these inefficiencies, recent approaches~\cite{huang2024rolora,lin2025duquant} avoid online rotation by merging rotational transformations ($R_1$) into consecutive linear layers. However, this introduces \textbf{architectural rigidity}, necessitating careful architectural adjustments (e.g., position encodings, residual, multi-head concat, layer norm). This design constraint hinders the generalization adaptability to emerging variants like multi-head latent attention blocks.

\paragraph{Bias-Centric Variants.}
Methods like Omniquant~\cite{shao2023omniquant} augment scaling with learnable bias terms:
\begin{equation}
\begin{split}
    \mathbf{Y} & = \mathbf{XW+B} = [\mathbf{(X-\delta)s^{-1}][sW]}+ (\mathbf{B} + \delta\mathbf{W})  \\ 
    & = \mathbf{\hat{X}\hat{W}} + \mathbf{\hat{B}} \approx \Delta_{\mathbf{\hat{X}}}\cdot(\mathbf{\hat{X}}_{int}\mathbf{\hat{W}}_{int})\Delta_{\mathbf{\hat{W}}} + \mathbf{\hat{B}},
\end{split}
\end{equation}
Where $\delta$ is a shift factor. However, bias terms do not resolve the fundamental coupling between weight and activation quantization, $\mathbf{\hat{W}}$ still depend on real-time $\mathbf{\hat{X}}$. Moreover, many LLMs (e.g., LLaMA, Phi-3) omit bias terms entirely, limiting generality. 

\section{Analysis of Outlier Spatial Stability Hypothesis}
\label{sec:OSSH}
\begin{figure}[t]
\centering
\includegraphics[width=0.8\linewidth]{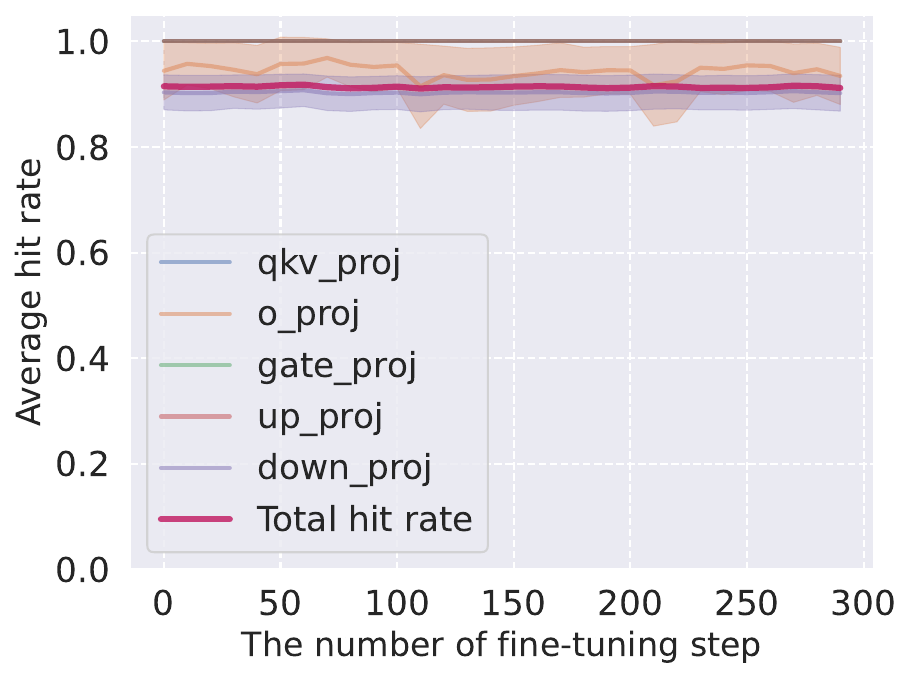}
\caption{Average hit rate of real-time vs. predefined outlier channel indices across layers in \textbf{LLaMA2-7B} during fine-tuning on OIG/Chip2.}
\label{fig:hitratellama}
\end{figure}

\begin{figure}[t]
\centering
\includegraphics[width=0.8\linewidth]{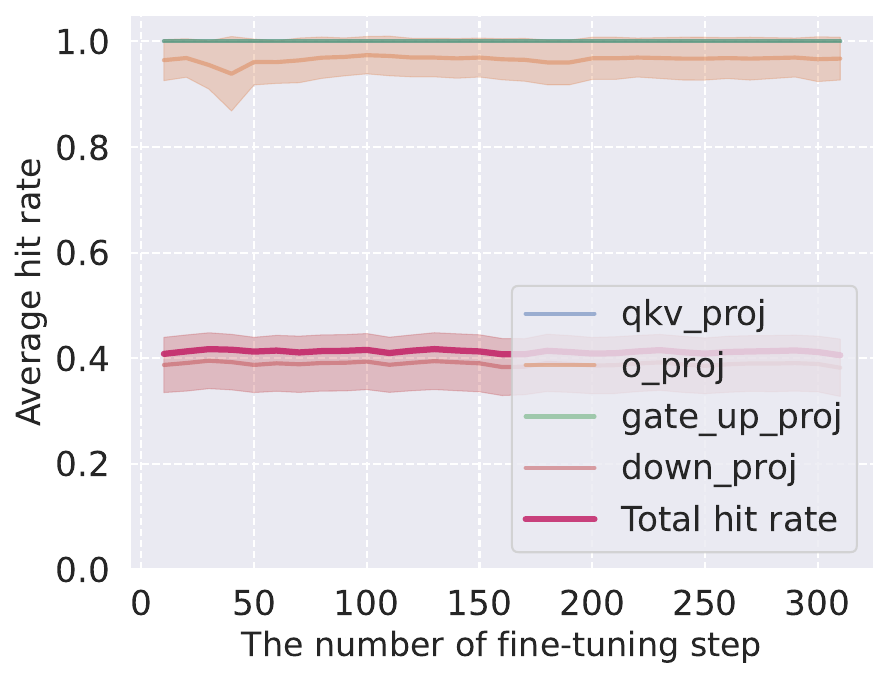}
\caption{Average hit rate of real-time vs. \textbf{uniformly} distributed predefined outlier channel indices across layers in Phi3-3.8B during fine-tuning on OIG/Chip2.}
\label{fig:hitratephi2}
\end{figure}

\paragraph{Empirical Validation.}
We validate the Outlier Spatial Stability Hypothesis (OSSH) by measuring the overlap between predefined outlier channels $O$ in fine-tuning iterations. And dynamically detected channels during fine-tuning. We analyze six core linear layers in transformer~\cite{waswani2017attention, devlin2018bert} of LLMs:
1. Attention projections: $q\_proj$, $k\_proj$, $v\_proj$. 2. Output projection: $o\_proj$. 3. FFN layers $up\_proj$ and  $down\_proj$. Notably, as previous works~\cite{lin2025duquant} indicate, $o\_proj$ and  $down\_proj$ exhibit higher outlier channel volatility due to input-sensitive activation patterns. To address this, we implement a non-uniform budget allocation: Stable layers $q\_proj$, $k\_proj$, $v\_proj$, $up\_proj$ have $0.03\%c_{in}$. Volatile layers $o\_proj$ have $4\%c_{in}$ and highly dynamic layers $down\_proj$ have $10\%c_{in}$. It should be noted that while the activations in $down\_proj$ and $o\_proj$ exhibit volatility, they still align with our OSSH, as the overall outlier channels set remain stable.
As shown in Figs.~\ref{fig:hitratellama} and \ref{fig:hitratephi2}, this strategy achieves $>90\%$ hit rates for LLaMA2-7B and Phi-3-3.8B on OIG/Chip2. In contrast, uniform budget allocation (Fig.~\ref{fig:hitratephi2}) reduces hit rates to $<50\%$ for volatile layers, confirming the necessity of layer-specific budget distribution.

\begin{figure}[t]
\centering
\includegraphics[width=0.8\linewidth]{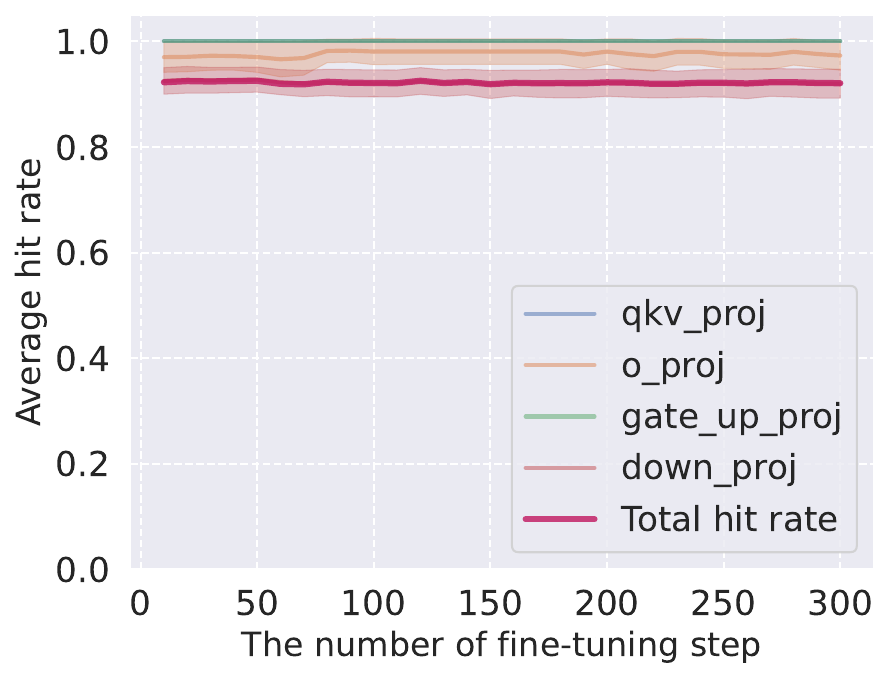}
\caption{Average hit rate of real-time vs. predefined outlier channel indices across layers in Phi-3-3.8B \textbf{during fine-tuning on GPQA.}}
\label{fig:hitrategpqa}
\end{figure}

\paragraph{Cross-Dataset Generalization.}
To test OSSH’s robustness, we evaluate Phi-3-3.8B on reasoning dataset GPQA using outlier channels calibrated on instruction-tuning OIG/Chip2. As shown in Fig.~\ref{fig:hitrategpqa}, hit rates remain >90\%, demonstrating hypothesis invariance across task domains.

\section{Outlier Distribution Shift Analysis}
\label{sec:shift}
\begin{figure}[t]
\centering
\includegraphics[width=0.85\linewidth]{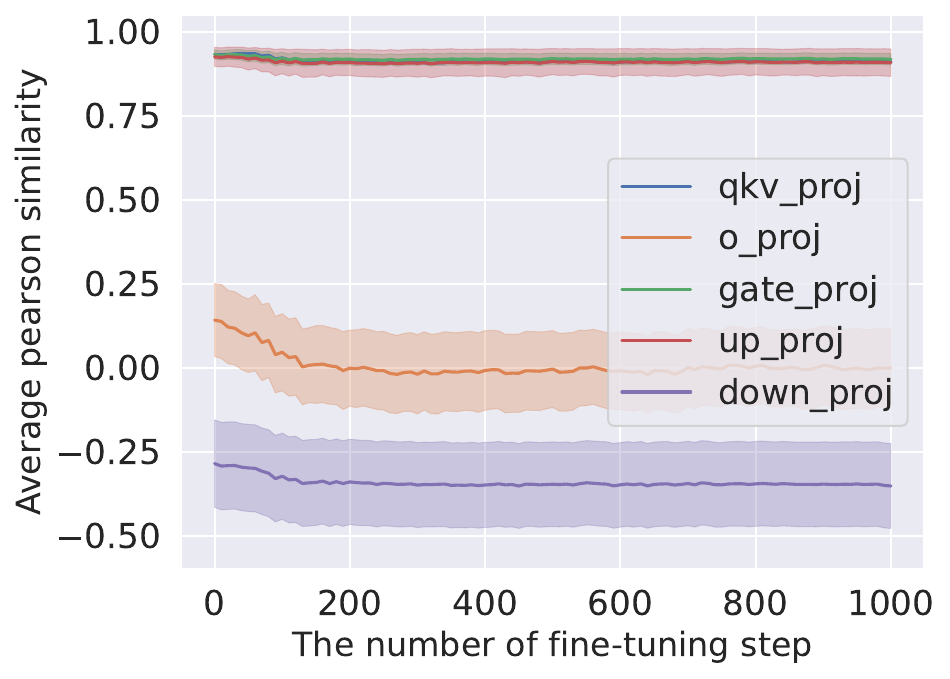}
\caption{Pearson similarity between static and dynamic scaling factors (top 1\%) across layers in LLaMA2-7B during fine-tuning on OIG/Chip2.}
\label{fig:scaling}
\end{figure}

Static scaling methods fail to adapt to activation distribution shifts, as evidenced by the declining similarity between predefined and real-time scaling factors (Fig.~\ref{fig:scaling}). In layers $down\_proj$, similarity drops to $-35\%$ after 1,000 iterations, explaining the accuracy degradation observed in prior work.

\section{More Experiment Analysis}

\subsection{The Impact of Long Context}
To explore The impact of Long Context, we investigated outlier stability on Phi-3 with a 32K input/output context. Using a 5\% budget, we pre-identified outlier channels via OIG/CHIP-2 and measured the average hit rate (\textit{i.e.,} whether an outlier appears in pre-identified outlier channels) during fine-tuning on the LongForm dataset. The results in Table~\ref{tb:32k} demonstrates the effectiveness of OSSH in by 90\% average hit rate in 32K context task.

\begin{table}[]
\centering
\begin{tabular}{l|l}
\hline
Layer & Average Hit rate \\ \hline
QKV\_proj & 100\% \\ \hline
gate\_up\_proj & 100\% \\ \hline
o\_proj & 98.3\% \\ \hline
down\_proj & 91.1\% \\ \hline
\end{tabular}
\caption{Average Hit rate for each type of layer in 32K context task.}
\label{tb:32k}
\end{table}

\subsection{The Impact of budget for outlier channels}
To show the impact of different budgets, we conduct experiments with different budget ratios. Since the current budget is sufficient, we focused on experiments exploring the impact of a lower budget. We evaluate budgets of 5\%, 3\%, 1\%, 0.1\%, and 0\% across different models and tasks. Specifically, we reduced the budget for $down\_proj$ and $o\_proj$ to obtain the overall budget of $3\%$ and $1\%$, and we set a uniform layer-wise budget of $0.1\%$ to achieve an overall budget of $0.1\%$. The results in the Table~\ref{tb:budgets} show that sensitivity decreases for long-text tasks (\textit{e.g.,} LAMBADA), smaller models (\textit{e.g.,} Phi-3).

\begin{table}[]
\centering
\resizebox{\linewidth}{!}{
\begin{tabular}{c|cc|cc}
\hline
\textbf{} & \multicolumn{2}{c|}{GPQA} & \multicolumn{2}{c}{LAMBADA} \\ \hline
 & llama2 7B & Phi 3 3.8B & llama2 7B & Phi 3 3.8B \\ \hline
5\% & 62.6 & 74.0 & 90.3 & 87.9 \\ \hline
3\% & 62.4 & 74.0 & 90.4 & 87.9 \\ \hline
1\% & 61.2 & 73.6 & 90.2 & 87.9 \\ \hline
0.1\% & 59.0 & 72.4 & 89.6 & 87.2 \\ \hline
0\% & 58.7 & 70.7 & 89.7 & 87.0 \\ \hline
\end{tabular}
}
\caption{Performance with different overall budgets.}
\label{tb:budgets}
\end{table}

\section{More Experimental Details}
\label{sec:exp_setting}
\paragraph{Dataset settings.} We set the number of fine-tuning epochs to 1 for instruction-tuning datasets (Alpaca-Finance, HH-RLHF, Self-Instruct, OIG/Chip2, and Oasst1) as well as long-text datasets (Longform and LAMBADA). For reasoning datasets (GPQA, MathQA, and MMLU-Pro), we set the number of fine-tuning epochs to 5. The prompts for instruction-tuning and long-text datasets follow their respective dataset settings, while for GPQA, MathQA, and MMLU-Pro, the prompt is:

    "\#Input Please select one of the following options: (A) \#Option1. (B) \#Option2. (C) \#Option3. (D) \#Option4."
    
and the reference text is formatted as :

\#Explanation. The answer is \#Correct.

And for MMLU-pro which does not provide sufficient explanation for training data, therefore, it left "\#Explanation" as blank.

\paragraph{Model and experimental settings.}We use Adam optimizer with learning rate as 2e-4 following previous work~\cite{dettmers2024qlora}. The rank of LoRA is 16, the alpha of LoRA is 16, and the LoRA dropout is 0.1. The number of virtual tokens in Prompt and P-tuning is set as 20.  We leverage bitsandbytes~\cite{dettmers2024qlora} for INT8 acceleration. We set $\gamma=0.2$ in the Equation~\ref{eq: momentum}.

\section{Quantization Granularity}
\label{sec:quant_gran}
Quantization has different levels of granularity related to different sizes of the quantization step size $\Delta_{\mathbf{X}}$ and $\Delta_{\mathbf{W}}$. The \textbf{\textit{per-tensor}} quantization uses one single quantization step for the entire matrix, i.e., $\Delta_{\mathbf{X}} \in \mathbb{R}$ and $\Delta_{\mathbf{W}} \in \mathbb{R}$, which can achieve fast quantization speed but high quantization loss. The \textbf{\textit{per-token}} and per-output-channel (\textbf{\textit{per-OC}}) quantization use different quantization step sizes for each token of activations and each output channel of weights, \textit{i.e.,} $\Delta_{\mathbf{X}} \in \mathbb{R}^{t}$ and $\Delta_{\mathbf{W}} \in \mathbb{R}^{c_{out}}$, which introduce lower quantization loss and higher computational overhead compared to per-tensor quantization. 
There is also a coarse-grained version of per-channel quantization called group-wise quantization (\textit{per-group})~\cite{yang2024gwq,jiang2024groupq}, which divides weight into different groups by classification and uses different quantization steps for different groups. However, per-group quantization requires specific hardware support for grouping operations. 
The per-input-channel (\textbf{\textit{per-IC}}) quantization~\cite{heo2023rethinking} using different quantization step sizes for each input channel, i.e., $\Delta_{\mathbf{X}} \in \mathbb{R}^{c_{in}}$ and $\Delta_{\mathbf{W}} \in \mathbb{R}^{c_{in}}$. However, per-IC quantization converts the MatMul into $\mathbf{Y} \approx \mathbf{X}_{int}\Delta_{\mathbf{X}} \Delta_{\mathbf{W}}\mathbf{W}_{int}$, which fails to achieve the integer MatMul for acceleration. \textit{Therefore, only per-tensor, per-token, and per-OC quantization methods can achieve both memory and computational efficiency on general hardware.}

\section{Broader Impact}
The deployment of large language models (LLMs) on personal and resource-constrained devices remains a key challenge due to the computational and memory demands of fine-tuning. This work introduces Quaff, a quantized parameter-efficient fine-tuning framework that enables efficient LLMs fine-tuning on consumer-grade hardware without full-precision weight storage. By leveraging the Outlier Spatial Stability Hypothesis (OSSH), Quaff facilitates hardware-friendly quantization, thereby bridging the gap between state-of-the-art language model capabilities and real-world accessibility.

This democratization of Quaff opens up significant societal and technological benefits in resource-constrained scenarios~\cite{huang2023distributed, huang2024fedmef}, such as mobile computing~\cite{forman1994challenges, imielinski1996mobile}, edge computing~\cite{cao2020overview, chen2024distributed} and cross-device federated learning~\cite{mcmahan2017communication, huang2025fedrts}. It empowers individuals, educators, small businesses, and developers in regions with limited computational resources to personalize and adapt LLMs for their specific needs, ranging from localized chatbots to domain-specific assistants without relying on cloud infrastructure. Furthermore, enabling on-device fine-tuning supports data privacy, as sensitive user data can remain local, reducing the risk of data leakage through centralized training.

However, with broader accessibility comes the potential for misuse. Easy personalization of LLMs could be exploited to fine-tune harmful behaviors or generate misinformation. We encourage the development of safeguards, such as differential privacy and fine-tuning auditing tools, to mitigate these risks.

Overall, Quaff advances the vision of inclusive and privacy-conscious AI by making powerful language technologies more accessible, efficient, and sustainable across a wide range of real-world environments.

\end{document}